\documentclass{iaesarticle}

\usepackage{amsmath, amsfonts, amssymb, float, fancyhdr}
\usepackage{algorithm} 
\usepackage{algpseudocode} 
\usepackage[figuresright]{rotating}
\usepackage{authblk, graphicx, indentfirst, lastpage, lipsum}
\usepackage{pifont}
\usepackage{multirow}
\usepackage[table,xcdraw]{xcolor}

\makeatletter
\renewcommand{\@biblabel}[1]{[#1]\hfill}
\makeatother
\usepackage{subfig, caption, epstopdf}
\usepackage[left=3cm, right=2.5cm, top=2.5cm, bottom=3.5cm, includehead, includefoot]{geometry}
\usepackage[justification=centering]{caption}
\usepackage{titlesec}
\usepackage{xcolor}
\titleformat{\section}
{\normalfont\normalsize\bfseries\uppercase}{\thesection}{1.7em}{}
\titleformat{\subsection}
{\normalfont\normalsize\bfseries}{\thesubsection}{.95em}{}
\titleformat{\subsubsection}
{\bfseries}{\thesubsubsection}{.2em}{}
\titlespacing*{\section}{0cm}{0.7cm}{0cm}
\usepackage{enumitem}
\usepackage[numbers,compress]{natbib}

\CopyrightLine[]{}{\textit{This is an open access article under the \textcolor{blue}{\underline{CC BY-SA}} license.}
\vspace{.5em}}

\author[1]{\bfseries Aminul Huq}
\author[2]{\bfseries Mst. Tasnim Pervin}

\affil[1]{Department of Computer Science and Engineering, Brac University, Bangladesh}
\affil[2]{Department of Computer Science and Technology, Tsinghua University, China}

\title{Adaptive Weight Assignment Scheme For Multi-task Learning}
\shorttitle{Adaptive Weight Assignment (Aminul Huq)}

\begin{document}
\setcounter{page}{1}

\setlength{\parindent}{1.27cm}

\pagestyle{fancy}
\fancyhfoffset{0cm}

\journalname{IAES International Journal of Artificial Intelligence (IJ-AI)}
\journalshortname{Int J Artif Intell}
\journalhomepage{http://ijai.iaescore.com}
\vol{x}
\no{x}
\months{March}
\years{202x}
\issn{2252-8938}
\DOI{10.11591}
\pagefirst{xx}
\pagelast{xx}

\maketitle

\hrule
\vspace{.1em}
\hrule
\vspace{.5em}
\noindent
\parbox[t][][s]{0.275\textwidth}{%
\textbf{Article Info}
\vspace{.5em}
\hrule
\vspace{.5em}
\begin{history}
\vspace{.5em}

Received Jun 12, 2018

Revised Aug 20, 2018

Accepted Aug 26, 2018

\vspace{.7em}
\end{history}
\vspace{.5em}
\hrule
\vspace{.5em}
\begin{keyword} 
\vspace{.5em}
Multi-task Learning \sep Adaptive Weight Assignment \sep Dynamic Weight Average\sep Uncertainty Weights\sep Deep Learning
\vspace{.5em}
\end{keyword}
\vspace{\fill}
}
\parbox{0.025\textwidth}{\hspace{0.5em}}
\parbox[t][][s]{0.7\textwidth}{%
\begin{abstract}
\vspace{.3em}
Deep learning based models are used regularly in every applications nowadays. Generally we train a single model on a single task. However, we can train multiple tasks on a single model under multi-task learning settings. This provides us many benefits like lesser training time, training a single model for multiple tasks, reducing overfitting, improving performances etc. To train a model in multi-task learning settings we need to sum the loss values from different tasks. In vanilla multi-task learning settings we assign equal weights but since not all tasks are of similar difficulty we need to allocate more weight to tasks which are more difficult. Also improper weight assignment reduces the performance of the model. We propose a simple weight assignment scheme in this paper which improves the performance of the model and puts more emphasis on difficult tasks. We tested our methods performance on both image and textual data and also compared performance against two popular weight assignment methods. Empirical results suggest that our proposed method achieves better results compared to other popular methods.

\end{abstract}
}
\parbox[l]{\textwidth}{%
\rule{0.275\textwidth}{0.5pt} \hspace{0.5cm} \hrulefill
\\
\emph{\textbf{Corresponding Author:}}
\vspace{.5em}\\
Aminul Huq,\\
Department of Computer Science and Engineering,\\
Brac University\\
66 Mohakhali, Dhaka-1212, Bangladesh.\\
Email: aminul.huq@bracu.ac.bd
}
\vspace{.5em}
\hrule
\vspace{.1em}
\hrule


\section{Introduction}
From the beginning of the last decade deep learning methods has been used vastly in various applications. The reach of it has exceeded tremendously not only in the field of computer science but also in electrical engineering, civil engineering, mechanical engineering and other fields as well. It is due to the fact that deep neural networks (DNN), have achieved human level competence in various applications like image classification \cite{DengDSLL009}, question answering \cite{RajpurkarZLL16}, lip reading \cite{assael2016lipnet}, video games \cite{chen2016evolution} etc. DNNs have the capability to find out complex and hidden features of the input data without any assistance. Previously these models were depended on hand crafted features \cite{DalalT05, Lowe04,MehrotraaR92,KhotanzadH90,pervin2017feature,huq2017combined}. 

Human beings have the capability to perform multiple tasks simultaneously without harming  performance of any tasks. Humans do this regularly and are able to decide which tasks can be done at the same time. That is why in recent years a lot of focus have been put into multi-task learning using DNN methods. Generally, a single model is devoted to performing a single task. 
However, performing multiple tasks increases the performance of the model, reduces training time and overfitting \cite{Caruana98}. Often we find small insufficient datasets for individual tasks but if the tasks are related somehow then we can use this shared information and build a large enough dataset which will reduce this problem. Currently in the field of mult-task learning, several research work is going on to create new DNN architectures for multi-task learning setting \cite{LiuJD19,MisraSGH16},  deciding which tasks should be learned together \cite{SunPFS20}, how to assign weights to the loss values \cite{KendallGC18,abs-1805-06334} etc. In this research work we focus on creating a dynamic weight assignment technique which will assign different weights to the loss values in each epoch during training. In our research work, we propose a new method for assigning weights to all loss values and test it against two datasets which are used in both image and text domain. The contributions of our research work are listed below. 

\begin{itemize}
    \item We propose an intuitive loss weighting scheme for multi-task learning. 
    \item We tested our method against both image and text domain by using two different dataset. We did this to ensure that our method performs well across all domains.
    \item We compared our method against two  popular weight assigning schemes for comparing the performance of our method. 
\end{itemize}
\section{Research Method}

In this section we will provide a discussion about previous research work performed in this field. Next, we will provide our proposed method.

\subsection{Literature Review}

One of the earliest papers on multi-task learning is provided by R. Caruana \cite{Caruana98}. In the manuscript, the author explored the idea of multi-task learning and showed it's effectiveness under different datasets. The author also explained how multi-task learning works and how it can be used in backpropagation. To train a DNN based on multi-task learning setting we need to consider which layers of network are shared among all the tasks and which layers are used for individual tasks. Previously, most of the research work has been focused on the concept of hard parameter sharing concept \cite{HuangFCY15,kokkinos2017ubernet,jou2016deep}. In this scenario, the user defines the shareable layers up to a particular point after which all layers are assigned per each task. There is also the concept of soft-parameter sharing where a single column exists for all the tasks in the network. A special mechanism is designed to share the parameter across all the network. Popular approaches for this method is Cross-stitch \cite{MisraSGH16}, Sluich \cite{RuderBAS19} etc. A new approach named Ada-share has been proposed recently where the model learns dynamically which layers to share for all tasks and which layers to be used for single tasks \cite{SunPFS20}. The authors also proposed a new loss function which ensures the compactness of the model as well as the performance of it. 

Weight assignment is a very crucial task in the field of multi-task learning. Previously weights either had equal values or some hand-tuned values which was assigned by the researchers \cite{kokkinos2017ubernet,EigenF15,SermanetEZMFL13}. However in scenarios where a large number of tasks existed for the multi-task learning model to perform,  such approaches fall short. A method based on uncertainty was proposed by \cite{KendallGC18}. Later a revised method of this approach was proposed by \cite{LiuJD19}. In this paper, the authors improved the previous uncertainty based method by adding a positive regularization term. Dynamic weight average method was proposed by \cite{LiuJD19}. In this method the authors calculated the relative change in loss values in previous two epochs and used softmax function on these values to get the weights. \cite{GongLSRPNKE19} performed a comparative study of different weight assigning scheme. However, they didn't study these methods in any domain other than images. Also, the dataset they used had only 2 tasks. 

\subsection{Adaptive Weight Assignment}

Our proposed method is simple and it takes into account of the loss value of each task in each epoch. Compared to other methods our method is easy to implement. Generally, in multi-task learning settings to train the model we need to sum up all the loss values with their weights and then perform backpropagation for updating the weights of the model. This summation of losses can be expressed as,

\begin{equation}
    \sum_{i=1,2,..n} W_{i}L_{i} = W_1L_1+W_2L_2+...+W_nL_n.
\end{equation}
Here, $W$ corresponds to the weight of the loss and $L$ represents the loss for each task. In vanilla multi-task learning setting all the weights are set to 1. However, we must keep in mind that all the tasks are not the same. Some are more difficult than others so we need to provide more weights on difficult tasks to improve performance of the overall multi-task learning system. That is why we propose Algorithm 1. 
\begin{algorithm}[ht]
	\textbf{Inputs:} Loss values ${L_1,L_2,..,L_n}$, total no. of tasks $n$
	\\
    \textbf{Outputs:} Total loss
	\begin{algorithmic}[1]
		\For {$t=1,2,\ldots n$}
		\State $TempLoss$ += $L_t$
	    \EndFor
	    \For {$t=1,2,\ldots n$}
		\State $weights_t$ = $L_t/TempLoss$
		\State $TotalLoss$ += $weights_t$ x $L_t$ x $n$  
	    \EndFor
	\end{algorithmic} 
\end{algorithm}
Our algorithm is based on the simple concept that difficult tasks will have more loss values than the easier ones. So we should put more emphasis or weights on those loss values while assigning less weights to the smaller loss values. What we do is take the sum of the loss values for each tasks and use it to figure out the ratio of how much a single tasks loss value contributes to the total loss. We multiply this value with the total number of tasks. Generally, in vanilla multi-task learning setting all loss values have equal weights 1. So the total weight is then n for n number of tasks. That is why we multiply our ratios with n. Finally, we use these  weights and using Eqn (1) compute the total loss for the multi-task learning model. Figure 1 provides a visual representation of the method.
\begin{figure}[ht]
    \centering
    {{\includegraphics[width=0.60\linewidth]{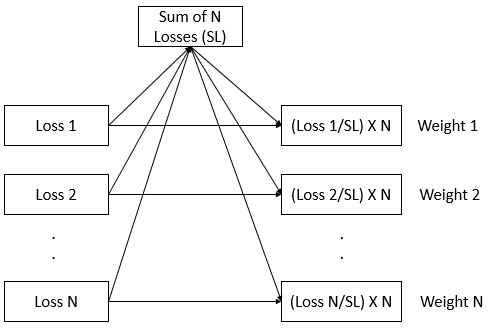} }}%
    \caption{Flow diagram of our proposed method.}%
    \label{fig:example1}%
\end{figure}
One of the important things about designing loss weighting schemes is that we need to ensure that these weight calculating methods should not take a lot time because it will increase the training time. Table 1 provides a chart about the time required to execute these schemes including our method. From the table we can see that though our method is not the fastest method to compute weights but it certainly is not the slowest. Also, the time difference between the quickest method and our method is very small. 
\begin{table}[ht]
\centering
\caption{Time required(s) for executing loss weighting schemes on CIFAR-100 and AGNews dataset.}
\label{tab:my-table1}
\begin{tabular}{ccc}
\hline
                & CIFAR-100 & AGNews \\ \hline
Re\_Uncertainty & 0.001     & 0.0004 \\ 
DWA             & 0.0004    & 0.0002 \\ 
Ours            & 0.0006    & 0.0003 \\ \hline
\end{tabular}
\end{table}
\section{Results and Discussion}

We will discuss about the dataset, experimental setup and results of the experiments in this section. 

\subsection{Dataset Description}
We used two different datasets in our experiment. They are CIFAR-100 \cite{krizhevsky2009learning} and AGNews \cite{ZhangZL15}. The formal one is image based and the later one is text based. Since these datasets are designed for single task learning we created artificial tasks for multi-task learning settings. We created 5 different tasks from CIFAR-100 and 2 tasks from AGNews dataset. All the tasks were created based on the original tasks labels and we grouped different labels together to form multiple tasks. The tasks were created to ensure that no class imbalance exists for all tasks. 

\subsection{Experimental Setup}
We used two different DNN models for our experiment. We used wide resnet-28-10 (WRN) \cite{ZagoruykoK16} for CIFAR-100 and a custom DNN for AGNews dataset. We split the final layer of the WRN model into 5 output layers for CIFAR-100 and 2 output layers for AGNews dataset. we trained WRN model for 100 epochs using SGD optimizer and set the learning rate to 0.001. We also used one cycle learning rate scheduler \cite{smith2019super}. In order to train the AGNews dataset we at first tokenize the dataset and create a vocabulary dictionary based on it. Then we perform embedding of the text which is going to be the input of the model. Our custom DNN consists of two fully connected layers. We trained this model using SGD optimizer. To ensure the effectiveness of our method, we compared our proposed method against two state-of-the-art methods namely dynamic weight average (DWA) and uncertainty method. We also compared against single task learning and vanilla multi-task setting.  

\subsection{Experimental Results}

We will discuss about the performance of our method against two datasets in this section. Table 2 and 3 represents the results of our overall experiment. We have plotted the testing loss curves for both CIFAR-100 and AGNews dataset in Figure 2.

In Table 2, we have the results on running experiments on CIFAR-100 dataset which is an image dataset. At the beginning we have results for all the five tasks in a single task learning settings. That is five different models were trained to get the results of these five tasks. Next under multi-task learning setting we trained four methods for these tasks. In vanilla multi-task learning we have assigned equal weights to each task for each epoch. Other methods Uncertainty, DWA and our method updates weights in each epoch. From this table we can see our proposed method out performs other methods in three out of five tasks. Also our method achieved second best performance in the rest of the two tasks. We can see that multi-task learning models performed better than STL models and also we needed to train only one single model for all five of these tasks. 

\begin{table}[ht]
\centering
\caption{Accuracy(\%)  comparison of different methods. Black (bold) marks the best score and red marks the second best score.}
\label{tab:my-table3}
\begin{tabular}{cccccc}
\hline
 &
  \begin{tabular}[c]{@{}c@{}}2 Class\\ Classification\end{tabular} &
  \begin{tabular}[c]{@{}c@{}}3 Class\\ Classification\end{tabular} &
  \begin{tabular}[c]{@{}c@{}}4 Class \\ Classification\end{tabular} &
  \begin{tabular}[c]{@{}c@{}}5 Class\\ Classification\end{tabular} &
  \begin{tabular}[c]{@{}c@{}}100 Class\\ Classification\end{tabular} \\ \hline
STL &
  74.52 &
  {\color[HTML]{FE0000} 75.70} &
  {\color[HTML]{FE0000} 74.02} &
  \textbf{72.81} &
  \textbf{76.56} \\ 
MTL - Vanilla &
  79.97 &
  74.36 &
  70.97 &
  67.95 &
  60.23 \\ 
MTL - Uncertainty &
  69.47 &
  59.52 &
  55.42 &
  50.21 &
  34.91 \\ 
MTL - DWA &
  {\color[HTML]{FE0000} 80.33} &
  74.57 &
  71.37 &
  68.41 &
  60.40 \\ 
MTL - Ours &
  \textbf{81.68} &
  \textbf{77.01} &
  \textbf{74.41} &
  {\color[HTML]{FE0000} 72.07} &
  {\color[HTML]{FE0000} 66.81} \\ \hline
\end{tabular}
\end{table}

\begin{figure}[ht]
    \centering
    \subfloat[\centering CIFAR-100]{{\includegraphics[width=0.46\linewidth]{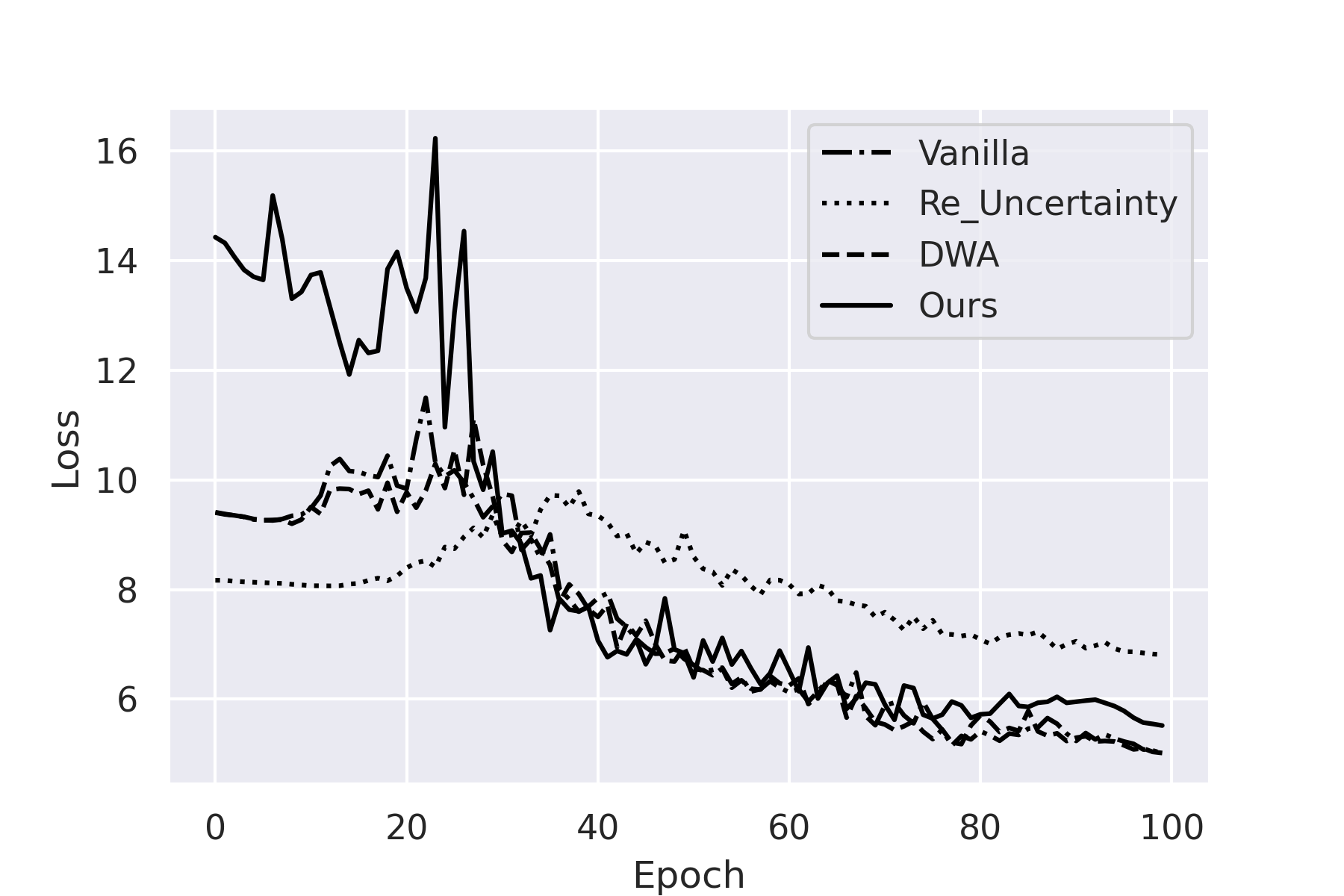} }}%
    \qquad
    \subfloat[\centering AGNews]{{\includegraphics[width=.46\linewidth]{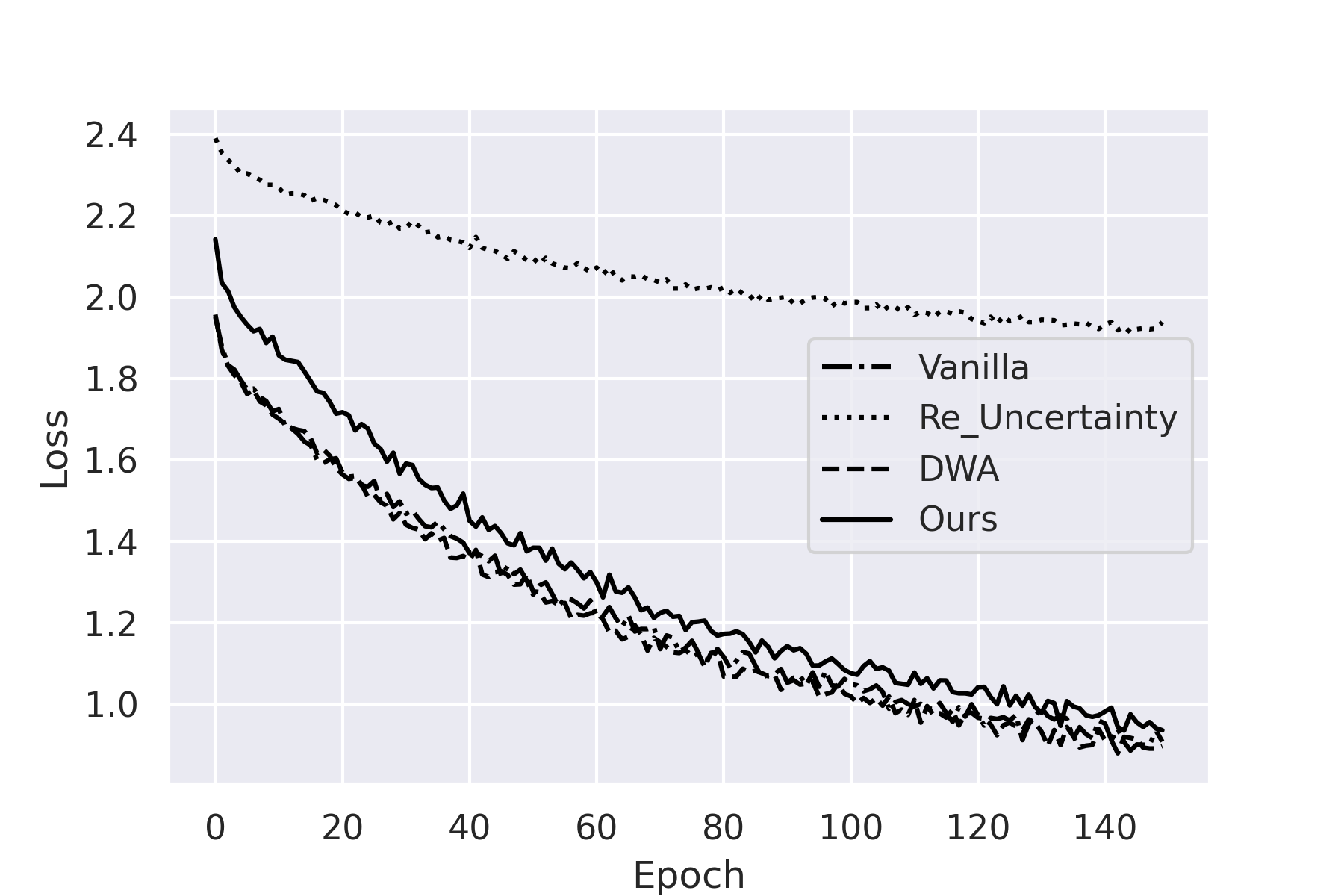} }}%
    \caption{Loss vs Epoch curve.}%
    \label{fig:example2}%
\end{figure}

We evaluate our methods performance on AGNews dataset which contains textual data. We have two tasks and at the beginning we train two individual models for these two tasks. After that we train four multi-task learning models with different weight assignment schemes.
We can observe from the table that our proposed method performs well under one task and achieves second best score in the other one. Compared to other popular methods we can see that our proposed method is performing much better. If we look closely at the values we will see that other methods fail to achieve the best results. In some cases these approaches even fail to attain better performance than single task learning approach. We believe this is due to the fact the model architecture has a big impact on the performance of multi-task learning settings. In our experiment we focused on uniform DNN architecture for evaluation but some tasks might need a few extra convolutional or fully connected layers. If we put further emphasis on the DNN architecture then the performance of our proposed method would definitely be better in both tasks.  We believe that a simpler approach should be taken while assigning weights. As this step is performed in each iteration, too much parameterized and complex approach mind hinder the performance of the model and increase time complexity.
\begin{table}[ht]
\centering
\caption{Accuracy(\%) comparison of different methods on AGNews dataset. Black (bold) marks the best score and red marks the second best score.}
\label{tab:my-table2}
\begin{tabular}{ccc}
\hline
 & \begin{tabular}[c]{@{}c@{}}2 Class\\ Classification\end{tabular} & \begin{tabular}[c]{@{}c@{}}4 Class \\ Classification\end{tabular} \\ \hline
STL               & 84.00                        & {\color[HTML]{333333} 79.13}          \\ 
MTL - Vanilla     & \textbf{86.57}               & {\color[HTML]{FE0000} 80.11}          \\ 
MTL - Uncertainty & 84.56                        & 75.94                                 \\ 
MTL - DWA         & {\color[HTML]{333333} 85.86} & 79.77                                 \\ 
MTL - Ours        & {\color[HTML]{FE0000} 86.02} & {\color[HTML]{333333} \textbf{81.18}} \\ \hline
\end{tabular}
\end{table}
\section{Conclusion}

Understanding and properly executing different hyper-parameters is extremely crucial in training a DNN model for the best results. Multi-task learning settings have the upper-hand on single task learning when it comes to amount of data needed, time to train the model, reducing overfitting and increasing model performance. In multi-task learning settings since not all tasks are of equal difficulties assigning weight to the loss values is important to put more emphasis on difficult task. In this paper, we propose a new weight assignment scheme which aids in improving the performance of the multi-task learning model. Our proposed method out-performs other state-of-the-art weight assigning schemes in both image and textual domain and boosts the performance of the model.





\bibliographystyle{IEEEtran}
\bibliography{refs.bib}

\begin{thebibliography}{10}
\providecommand{\url}[1]{#1}
\csname url@samestyle\endcsname
\providecommand{\newblock}{\relax}
\providecommand{\bibinfo}[2]{#2}
\providecommand{\BIBentrySTDinterwordspacing}{\spaceskip=0pt\relax}
\providecommand{\BIBentryALTinterwordstretchfactor}{4}
\providecommand{\BIBentryALTinterwordspacing}{\spaceskip=\fontdimen2\font plus
\BIBentryALTinterwordstretchfactor\fontdimen3\font minus
  \fontdimen4\font\relax}
\providecommand{\BIBforeignlanguage}[2]{{%
\expandafter\ifx\csname l@#1\endcsname\relax
\typeout{** WARNING: IEEEtran.bst: No hyphenation pattern has been}%
\typeout{** loaded for the language `#1'. Using the pattern for}%
\typeout{** the default language instead.}%
\else
\language=\csname l@#1\endcsname
\fi
#2}}
\providecommand{\BIBdecl}{\relax}
\BIBdecl

\bibitem{DengDSLL009}
J.~Deng, W.~Dong, R.~Socher, L.~Li, K.~Li, and F.~Li, ``Imagenet: {A}
  large-scale hierarchical image database,'' in \emph{{IEEE} Computer Society
  Conference on Computer Vision and Pattern Recognition {(CVPR})}, 2009, pp.
  248--255.

\bibitem{RajpurkarZLL16}
P.~Rajpurkar, J.~Zhang, K.~Lopyrev, and P.~Liang, ``Squad: 100, 000+ questions
  for machine comprehension of text,'' in \emph{Proceedings of the 2016
  Conference on Empirical Methods in Natural Language Processing, {EMNLP}},
  2016, pp. 2383--2392.

\bibitem{assael2016lipnet}
Y.~M. Assael, B.~Shillingford, S.~Whiteson, and N.~De~Freitas, ``Lipnet:
  End-to-end sentence-level lipreading,'' \emph{arXiv preprint
  arXiv:1611.01599}, 2016.

\bibitem{chen2016evolution}
J.~X. Chen, ``The evolution of computing: Alphago,'' \emph{Computing in Science
  \& Engineering}, vol.~18, no.~4, pp. 4--7, 2016.

\bibitem{DalalT05}
N.~Dalal and B.~Triggs, ``Histograms of oriented gradients for human
  detection,'' in \emph{Computer Vision and Pattern Recognition (CVPR)}.\hskip
  1em plus 0.5em minus 0.4em\relax {IEEE} Computer Society, 2005, pp. 886--893.

\bibitem{Lowe04}
D.~G. Lowe, ``Distinctive image features from scale-invariant keypoints,''
  \emph{International Journal on Computer Vision}, vol.~60, no.~2, pp. 91--110,
  2004.

\bibitem{MehrotraaR92}
R.~Mehrotra, K.~R. Namuduri, and N.~Ranganathan, ``Gabor filter-based edge
  detection,'' \emph{Pattern Recognition}, vol.~25, no.~12, pp. 1479--1494,
  1992.

\bibitem{KhotanzadH90}
A.~Khotanzad and Y.~H. Hong, ``Invariant image recognition by zernike
  moments,'' \emph{IEEE Transactions on Pattern Analysis and Machine
  Intelligence}, vol.~12, no.~5, pp. 489--497, 1990.

\bibitem{pervin2017feature}
M.~T. Pervin, S.~Afroge, and A.~Huq, ``A feature fusion based optical character
  recognition of bangla characters using support vector machine,'' in \emph{3rd
  international conference on electrical information and communication
  technology (EICT)}.\hskip 1em plus 0.5em minus 0.4em\relax IEEE, 2017, pp.
  1--6.

\bibitem{huq2017combined}
A.~Huq, S.~Afroge, and M.~T. Pervin, ``Combined zernike moments, binary pixel
  and histogram of oriented gradients feature extraction technique for
  recognizing hand written bangla characters,'' in \emph{3rd International
  Conference on Electrical Information and Communication Technology
  (EICT)}.\hskip 1em plus 0.5em minus 0.4em\relax IEEE, 2017, pp. 1--5.

\bibitem{Caruana98}
R.~Caruana, ``Multitask learning,'' in \emph{Learning to Learn}, S.~Thrun and
  L.~Y. Pratt, Eds.\hskip 1em plus 0.5em minus 0.4em\relax Springer, 1998, pp.
  95--133.

\bibitem{LiuJD19}
S.~Liu, E.~Johns, and A.~J. Davison, ``End-to-end multi-task learning with
  attention,'' in \emph{{IEEE} Conference on Computer Vision and Pattern
  Recognition, {CVPR}}, 2019, pp. 1871--1880.

\bibitem{MisraSGH16}
I.~Misra, A.~Shrivastava, A.~Gupta, and M.~Hebert, ``Cross-stitch networks for
  multi-task learning,'' in \emph{{IEEE} Conference on Computer Vision and
  Pattern Recognition, {CVPR}}, 2016, pp. 3994--4003.

\bibitem{SunPFS20}
X.~Sun, R.~Panda, R.~Feris, and K.~Saenko, ``Adashare: Learning what to share
  for efficient deep multi-task learning,'' in \emph{Annual Conference on
  Neural Information Processing Systems, NeurIPS}, 2020.

\bibitem{KendallGC18}
A.~Kendall, Y.~Gal, and R.~Cipolla, ``Multi-task learning using uncertainty to
  weigh losses for scene geometry and semantics,'' in \emph{{IEEE} Conference
  on Computer Vision and Pattern Recognition, {CVPR}}, 2018, pp. 7482--7491.

\bibitem{abs-1805-06334}
\BIBentryALTinterwordspacing
L.~Liebel and M.~K{\"{o}}rner, ``Auxiliary tasks in multi-task learning,''
  \emph{CoRR}, vol. abs/1805.06334, 2018. [Online]. Available:
  \url{http://arxiv.org/abs/1805.06334}
\BIBentrySTDinterwordspacing

\bibitem{HuangFCY15}
J.~Huang, R.~S. Feris, Q.~Chen, and S.~Yan, ``Cross-domain image retrieval with
  a dual attribute-aware ranking network,'' in \emph{{IEEE} International
  Conference on Computer Vision, {ICCV}}, 2015, pp. 1062--1070.

\bibitem{kokkinos2017ubernet}
I.~Kokkinos, ``Ubernet: Training a universal convolutional neural network for
  low-, mid-, and high-level vision using diverse datasets and limited
  memory,'' in \emph{{IEEE} Conference on Computer Vision and Pattern
  Recognition, {CVPR}}, 2017, pp. 6129--6138.

\bibitem{jou2016deep}
B.~Jou and S.-F. Chang, ``Deep cross residual learning for multitask visual
  recognition,'' in \emph{24th ACM international conference on Multimedia},
  2016, pp. 998--1007.

\bibitem{RuderBAS19}
S.~Ruder, J.~Bingel, I.~Augenstein, and A.~S{\o}gaard, ``Latent multi-task
  architecture learning,'' in \emph{The Thirty-Third {AAAI} Conference on
  Artificial Intelligence, {AAAI}}, 2019, pp. 4822--4829.

\bibitem{EigenF15}
D.~Eigen and R.~Fergus, ``Predicting depth, surface normals and semantic labels
  with a common multi-scale convolutional architecture,'' in \emph{{IEEE}
  International Conference on Computer Vision, {ICCV}}, 2015, pp. 2650--2658.

\bibitem{SermanetEZMFL13}
P.~Sermanet, D.~Eigen, X.~Zhang, M.~Mathieu, R.~Fergus, and Y.~LeCun,
  ``Overfeat: Integrated recognition, localization and detection using
  convolutional networks,'' in \emph{2nd International Conference on Learning
  Representations, {ICLR},}, 2014.

\bibitem{GongLSRPNKE19}
T.~Gong, T.~Lee, C.~Stephenson, V.~Renduchintala, S.~Padhy, A.~Ndirango,
  G.~Keskin, and O.~H. Elibol, ``A comparison of loss weighting strategies for
  multi task learning in deep neural networks,'' \emph{{IEEE} Access}, vol.~7,
  pp. 141\,627--141\,632, 2019.

\bibitem{krizhevsky2009learning}
A.~Krizhevsky, G.~Hinton \emph{et~al.}, ``Learning multiple layers of features
  from tiny images,'' 2009.

\bibitem{ZhangZL15}
X.~Zhang, J.~J. Zhao, and Y.~LeCun, ``Character-level convolutional networks
  for text classification,'' in \emph{Annual Conference on Neural Information
  Processing Systems, NeurIPS}, 2015, pp. 649--657.

\bibitem{ZagoruykoK16}
S.~Zagoruyko and N.~Komodakis, ``Wide residual networks,'' in \emph{British
  Machine Vision Conference, {BMVC}}, 2016.

\bibitem{smith2019super}
L.~N. Smith and N.~Topin, ``Super-convergence: Very fast training of neural
  networks using large learning rates,'' in \emph{Artificial Intelligence and
  Machine Learning for Multi-Domain Operations Applications}, vol. 11006, 2019,
  p. 1100612.

\end{thebibliography}



\section*{BIOGRAPHIES OF AUTHORS} 
\vspace{-.7em} 

\begin{biography}[{\includegraphics[width=2.5cm,height=4cm,clip,keepaspectratio]{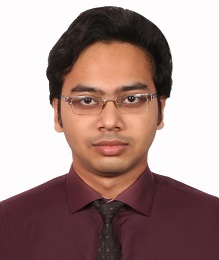}}]
\textbf{Aminul Huq} 
is a lecturer at Brac University. He received his Master's in Computer Science and Technology from Tsinghua University in 2021. He completed his Bachelor's from Rajshahi University of Engineering \& Technology in 2017. His research interest lies in the field of Multi-task Learning and Adversarial Machine Learning. Email : aminul.huq@bracu.ac.bd 

\end{biography}

\begin{biography}[{\includegraphics[width=2.5cm,height=4cm,clip,keepaspectratio]{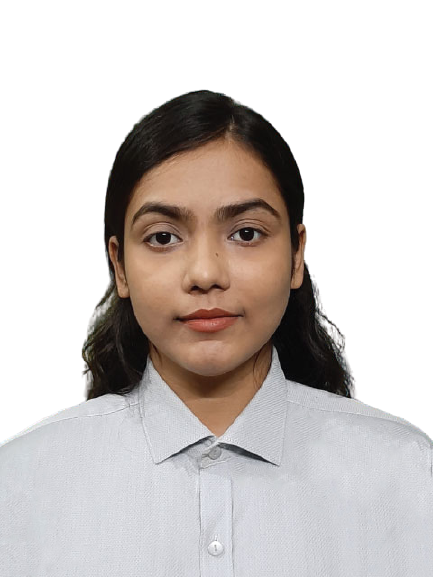}}]
\textbf{Mst. Tasnim Pervin} 
has completed her Master's in Computer Science and Technology from Tsinghua University in 2021. She completed her Bachelor's from Rajshahi University of Engineering \& Technology in 2017. Her research interest lies in the field of Medical Image Analysis, Adversarial Machine Learning and Domain Adaptation. Email : pervinmt10@mails.tsinghua.edu.cn

\end{biography}

\end{document}